# Effective Computer Model For Recognizing Nationality From Frontal Image

Bat-Erdene.B
Information and Communication Management School
The University of the Humanities
Ulaanbaatar, Mongolia
e-mail: basubaer@gmail.com

Ganbat.Ts
School of Information and Communication Technology
The Mongolian University of Science and Technology
Ulaanbaatar, Mongolia
e-mail: ganbat_tsend@yahoo.com

*Abstract*—We are introducing new effective computer model for extracting nationality from frontal image candidate using face part color, size and distances based on deep research. Determining face part size, color, and distances is depending on a variety of factors including image quality, lighting condition, rotation angle, occlusion and facial emotion. Therefore, first we need to detect a face on the image then convert an image into the real input. After that, we can determine image candidate's gender, face shape, key points and face parts. Finally, we will return the result, based on the comparison of sizes and distances with the sample's measurement table database. While we were measuring samples, there were big differences between images by their gender and face shapes. Input images must be the frontal face image that has smooth lighting and does not have any rotation angle. The model can be used in military, police, defense, healthcare, and technology sectors. Finally, Computer can distinguish nationality from the face image.

## I. Introduction

Face recognition provides a great opportunity to creating a useful application to identify humans for the security and immigration offices in some countries. One of the major problems in modeling face recognition and processing is to find out a solution of representing faces that helps for the variety of tasks typical of human performance. The human face is a complex visual pattern that includes general categorical information as well as eccentric, identity specific, primary information. By this categorical information, we mean that some aspects of a face are not specific to that particular face but are shared by subsets of faces. These aspects can be used to appoint both unfamiliar and familiar faces to general semantic groups such as region or nationality. Traditionally, computational models of face recognition represent faces in terms of geometric identifications that contain distances, angles, and areas between base features such as eyes, nose, mouth or chin. Nowadays there are some effective algorithms for detecting face candidate region. Even though some only works with the manual measurements, others detecting face region [1]. About the detection of face candidate nationality, especially research on recognizing nationality have not been researched yet. To make a decision to this problem, first we have collected a thousand frontal image from every country (Mongolian, Japanese, Chinese, and Korean) to create our sampling. Then we have measured all of them to establish our measurement table database [2]. The measurement provides great chances to processing images with the database. For further development, we are working on creating a full database that grouped through the common facial features.

## II. System Model

First of all, our system detects the face from input image [5]. After that, we can determine image candidate's gender, face shape, key points and face parts. Then, we will return the result, based on the comparison of sizes and distances with the sample's measurement table database with the support vector machine (SVM) and active shape model (ASM).

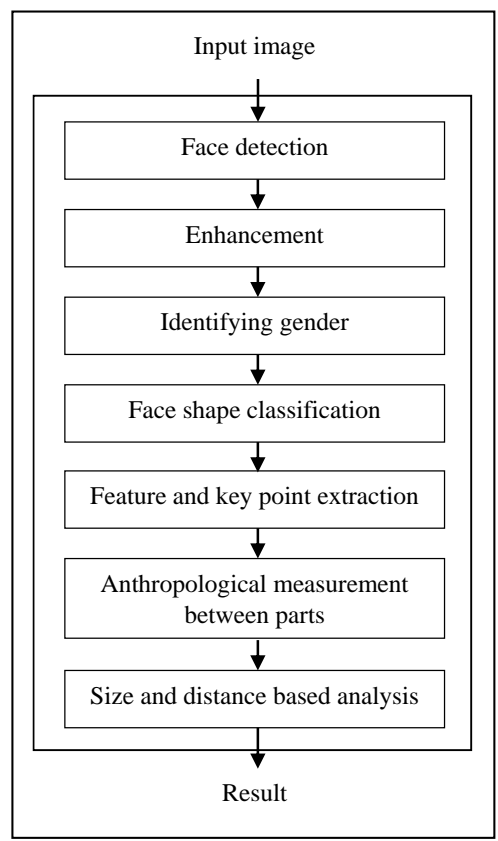

Figure 1. Flowchart of the system model

## A. Face detection

Face detection is a process to divide an image into the two pieces: one contains a face and the other non-face image. Final detection result has a functional relation to the process time and overall result. A variety of methods has been proposed, and some have been brought into exploitation in the real system.

In general, those methods can be divided into two major sections: the first section is based on all kind of face parts, which means the final result comes from the integration of several detection results. The second considers a face as a single detection input and facial features are extracted from the entire face region. We are going to use methods in both sections to achieve higher performance and better accuracy. A cascade is used to reduce image processing time by focusing attention on the more interesting regions of the image. For example, the flat regions of an image, clearly do not contain faces and can be quickly discarded by use of a template consisting of only a small number of features. Such a scheme has the potential to greatly improve the speed of the detector but still allow a large number of features to be evaluated on highly textured regions that may contain faces.

## B. Enhancement

The better performance and accuracy are related to input lighting, rotation angle, facial expression, and image quality. Hence, in this stage image size, quality lighting and rotation angle are normalized until real input [3]. In our system, we used edge detection to improve the contrast of the real input.

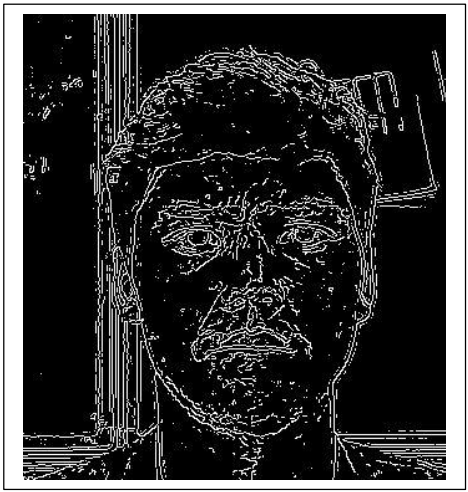

Figure 2.  Edge detection of face

## C. Identifying gender

While we were taking a measure from the sample images, there was interval exception from the image candidate's gender. In real biological systems, especially in an anthropological and morphological analysis calculator measurement must have been done with the gender classification [6][7]. The major difficulty of representing faces as a set of features is that it presumes some a priori knowledge about what are the main features and what are the relationships between them that are essential to the task. It is very difficult to find out a set of features helpful in discriminating accurately between male and female faces. First we try to categorize with their facial feature and landmark distances. Our experiment shows that no simple set of features can predict the gender of the faces accurately. Thus, we did a number of experiments for the gender categorization. Finally, we selected Principal component analysis (PCA) and eigenvector to group faces into the male and female. The reason for this, the Eigen face technique relies on information theory and take advantage of PCA, the method of decreasing dimensionality while preserving the variance of a data set, to recognize facial gender. More deeply, the principal components used in the Eigen face technique are the eigenvector of the covariance matrix of faces, while every face is a point in n space where n is the number of pixel in each image [10].

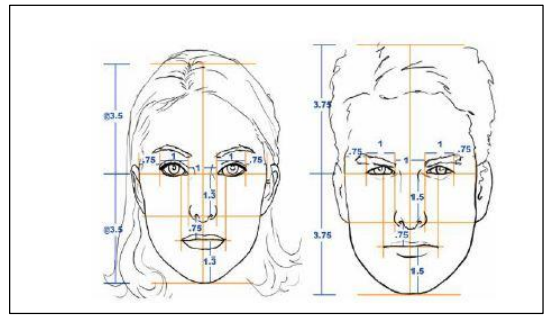

Figure 3.  Face proportions [11]

## D. Face shape classification

Human faces are categorized into seven geometric shapes. Using Fig. 4's base shapes (from right oval, square, round, triangle, heart, oblong and diamond) we will identify the shapes from the facial image [4]. Our research shows that Most Japanese people have a longer and oval face with wider eyes. Most of the Korean people have a flatter face with square cheekbones and with smaller eyes. Chinese people have rounder face with thinner cheekbones and with middle sized eyes. Mongolian people have the heart face with wider cheekbones. There are the variety of influences in a human face. Such as nationality genetic, weather, living environment, and eating food. Mongolian people eat a meal every day, especially they eat more meal in Mongolian national holidays. That was the big influence of their wider cheekbone. In our system, we used active appearance model (AAM) and ASM to recognize face shapes.

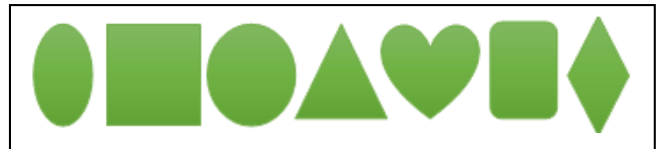

Figure 4.  Face shape categories

## E. Feature and key point extraction

While we are working on real-time video, facial expression, and occlusions involve simultaneous changes of facial features on multiple face parts. In most systems, facial feature extraction includes locating the position and shape of the eyebrows, eyes, eyelids, mouth, wrinkles, nose, and extracting features related to them in a still image of the human face. For a particular facial activity, there is a subset of the facial features that are the most informative and maximally reduces the ambiguity of classification. Therefore, we actively select facial visual cues to achieve a desirable result in a timely and efficient manner while reducing the ambiguity of classification to a minimum. In our model, there are two types of facial features: base and extra features. The base facial features are the shapes and location of eyes, nose, mouth and lips. The extra features we have to detect are forehead, mandible, eyebrows, eyelid, chin and ear. We have measured all of these features by the color, size, shape and distance from the neighbors.

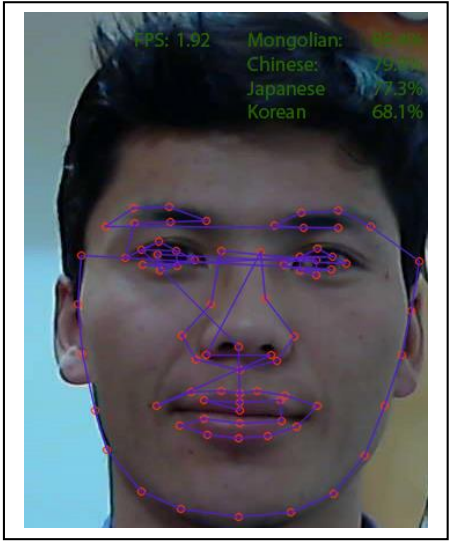

Figure 5. Face detection key points

## F. Anthropological measurement between parts [7]

After feature and key point extraction, we calculated the distance from the neighbors using the landmark and edge detection [8]. Then we will have full measurement collection from the image. According to the Fig. 6, there are fourteen measurements in anthropological analysis and ten measurements in morphological analysis from frontal face [9].

We chose suitable key points from that analysis and we added some extra points and measurements to our analysis. After this process, we convert an input image into the point vectors. Then we transfer vector collection to prediction section of support vector machine.

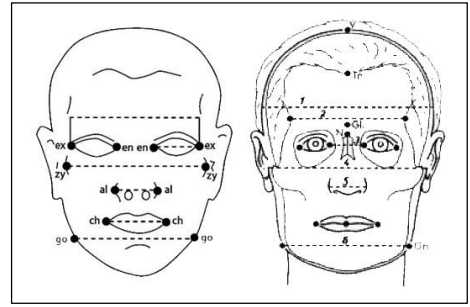

Figure 6. Antrophological and Morphological analysis points

## G. Size and distance based analysis

Above-mentioned measurements are created on frontal face images by the following values [7].

TABLE I. SIZE AND DISTANCE TABLE

| № | MEASUREMENT UNITS | |
|---|---|---|
| | *Size* | *Distance* |
| 1 | Head width, height | Eye distance |
| 2 | Forehead width, height | Eyebrow distance |
| 3 | Face width | Ear distance |
| 4 | Nose width, height | Chin to lip distance |
| 5 | Ear height | Lip to nose distance |
| 6 | Lip width | Nose to eye distance |
| 7 | Eye width, height | |
| 8 | Eyebrow width | |
| 9 | Chin width | |

In this stage, we predict and calculate nationality percent of input point vectors. Our facing problem was how to check and validate our prediction according to Fig. 7. To solve this problem, we used mean square error and regression analysis. Using those methods, we did error control and prediction validation.

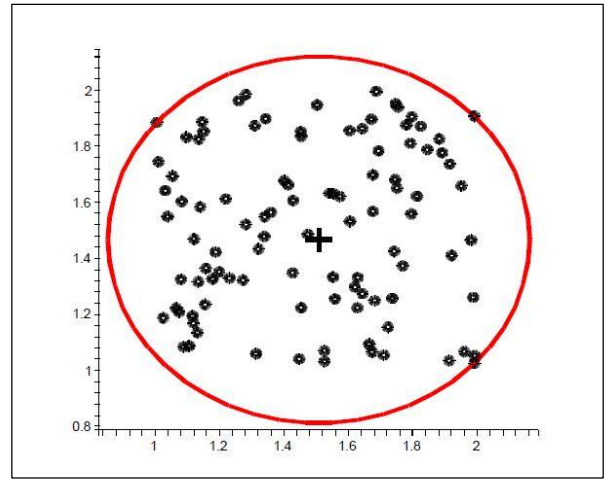

Figure 7. SVM prediction ot the system

**Overflow of Proposed System:**
Step 1: Video Capture from real-time video with openCV
Step 2: Face detection with classifier
Step 3: Do enhancement with edge detection until real input
Step 4: Do gender classification with PCA and Eigen face
Step 5: Classify face shapes with AAM and ASM
Step 6: Extract feature and key point with openCV, add to vectors collection
Step 7: Calculate measurements with Euclidean distance, add to vectors collection
Step 8: SVM based analysis using vector collection
Step 9: Return result percent
Step 10: if highest percent is more than 86.4%, save to database with the suitable category.

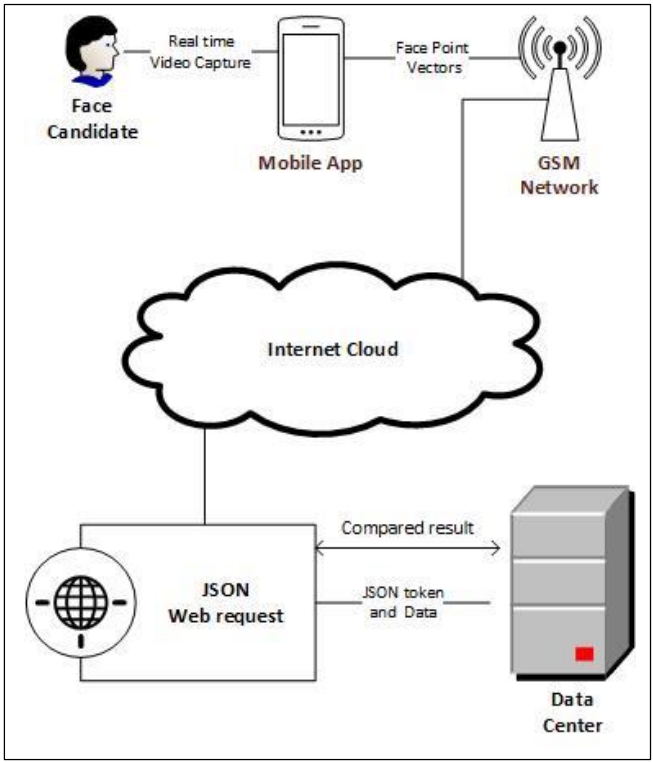

Figure 8. Proposed system diagram

### III. EXPERIMENTS AND RESULTS

First of all, we made facial point collector software and national model creator script on visual studio C# environment with the EmguCV to create a sample database. Then we saved our collected points to the Microsoft SQL server and we converted all of the data collection into the four model file. After that, we did experiments to choose the best fit method for our system by their true rate, speed, and rationality. Then we developed nationality recognition software on four platforms. Eventually, we chose an android platform for the real application.

While we are doing experiments, all of the methods have advantages and disadvantages. But we mixed them together to get great results.

TABLE II. SIZE AND DISTANCE TABLE

| № | EXPERIMENTS ||||| 
|---|---|---|---|---|---|
| | *Platform* | *Language* | *Library* | *Methods* | *Rate* |
| 1 | QT | C++ | OpenCV | Haar-Cascade classifier, support vector machine, AAM, ASM | 81.5% |
| 2 | Visual Studio | C# | EmguCV | | 78.3% |
| 3 | Matlab | Matlab | Facelib | | 80.7% |
| 4 | Android | Java | openCV, libsvm, stasm | | 86.4% |

We made an android application that can distinguish people nationality by their single frontal image to collect more data into our database.

Our system diagram is shown in Fig. 8.

### CONCLUSIONS

In this paper, we have presented a novel computer model for extracting nationality from frontal image candidate. First, a number of faces should be known before the process. Hence, we have collected and created our facial image database by the four countries.

To distinguish nationality, we have done two kind of feature extractions. It gives the great opportunity to reduce similarities. The proposed model is effective for frontal face images with the normal lighting condition and with the normal rotation angle. Using this model, we can recognize nationality from frontal image candidate on the machine automatically.

There are some existing challenges in using nationality detection to create high qualified applications for the military, police, defense, and immigration office. An additionally we can use our system to expand real business model for electronic marketing. In the future, we will work on creating a full database that grouped through the common facial features from all of the countries for creating a corporative information system that enables internet of human.